\documentclass[conference]{IEEEtran}
\IEEEoverridecommandlockouts
\usepackage{amsmath,amssymb,amsfonts}
\usepackage{algorithmic}
\usepackage{graphicx}
\usepackage{textcomp}
\usepackage{xcolor}
\usepackage[numbers,sort&compress]{natbib}
\usepackage{siunitx}
\usepackage{url}
\usepackage{multirow}
\usepackage{enumitem}
\usepackage{subcaption}
\usepackage{booktabs}
\def\BibTeX{{\rm B\kern-.05em{\sc i\kern-.025em b}\kern-.08em
   T\kern-.1667em\lower.7ex\hbox{E}\kern-.125emX}}
\begin{document}
\title{LEO: Graph Attention Network based Hybrid Multi\\
Sensor Extended Object Fusion and Tracking for Autonomous Driving Applications\\
% {\footnotesize \textsuperscript{*}Note: Sub-titles are not captured for https://ieeexplore.ieee.org  and
\thanks{This work is a result of the joint research project STADT:up (19A22006O). The project is supported by the German Federal Ministry for Economic Affairs and Energy (BMWE), based on a decision of the German Bundestag. The author is solely responsible for the content of this publication. We would also like to thank Mr. Lukas Ostendorf, IKA RWTH Aachen University for his valuable technical insights, discussions, and feedback on this research.}
}

\author{\IEEEauthorblockN{Mayank Mayank\IEEEauthorrefmark{1}, Bharanidhar Duraisamy\IEEEauthorrefmark{1}, Florian Geiss\IEEEauthorrefmark{1}} \\
\IEEEauthorblockA{\IEEEauthorrefmark{1} Research and Development, Mercedes-Benz AG, Germany, Email: [firstname].[lastname]@mercedes-benz.com}
% \and
% \and
% \and
}

\maketitle

\begin{abstract}
Accurate shape and trajectory estimation of dynamic objects is essential for reliable automated driving. Classical Bayesian extended-object models offer theoretical robustness and efficiency but depend on completeness of a-priori and update-likelihood functions, while deep learning methods bring adaptability at the cost of dense annotations and high compute. We bridge these strengths with LEO (Learned Extension of Objects), a spatio-temporal Graph Attention Network that fuses multi-modal production-grade sensor tracks to learn adaptive fusion weights, ensure temporal consistency, and represent multi-scale shapes. Using a task-specific parallelogram ground-truth formulation, LEO models complex geometries (e.g. articulated trucks and trailers) and generalizes across sensor types, configurations, object classes, and regions, remaining robust for challenging and long-range targets. Evaluations on the Mercedes‑Benz DRIVE PILOT SAE L3 dataset demonstrate real-time computational efficiency suitable for production systems; additional validation on public datasets such as View of Delft (VoD) further confirms cross-dataset generalization.
\end{abstract}


\section{Introduction}
Autonomous Driving (AD) is a key enabler of safer, more efficient, and inclusive transportation systems. Human error accounts for nearly 94\% of severe traffic accidents, highlighting the potential of Autonomous Vehicles (AVs) to improve safety through consistent, rule-based decision making and enhanced situational awareness \cite{singh2015critical}. Beyond safety, AD promises improved mobility for elderly and disabled users, reduced congestion through coordinated traffic flow, and lower operating costs via energy-efficient driving and shared mobility concepts \cite{fagnant2015preparing,yurtsever2020survey}. These benefits have driven significant academic and industrial investment, establishing AD as a central pillar of future intelligent transportation systems \cite{badue2021self}.

\begin{figure*}[t]
    \centering
    % First subfigure
    \begin{subfigure}[t]{0.40\textwidth}
        \centering
        \includegraphics[width=\textwidth, trim=285 60 50 160, clip]{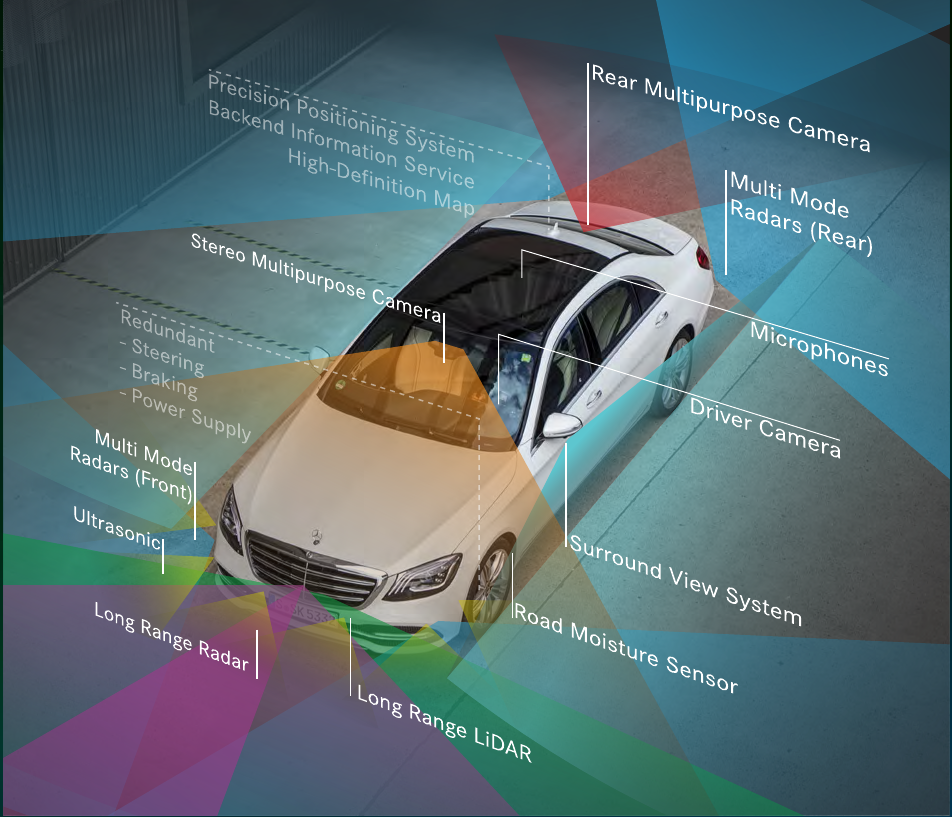}
        \caption{Mercedes-Benz EQS sensors.}
        \label{fig:MB_DrivePilot}
    \end{subfigure}
    \hspace{1cm}
    % Second subfigure
    \begin{subfigure}[t]{0.40\textwidth}
        \centering
        \includegraphics[width=\linewidth]{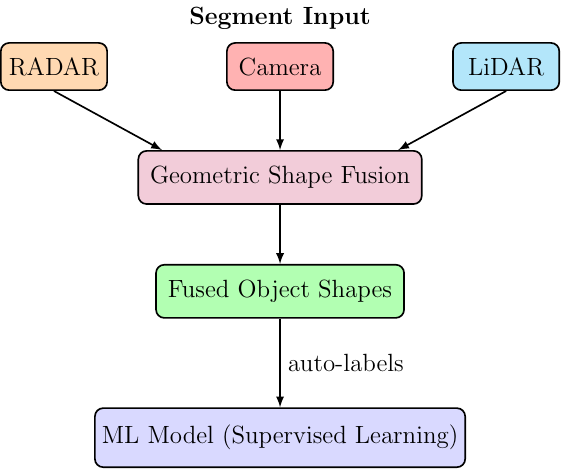}
        \caption{Supervised learning using labels from geometric method.}
        \label{fig:autolabelling}
    \end{subfigure}

    \caption{Mercedes-Benz EQS sensors used by DRIVE PILOT \cite{MB-DrivePilot2023-1} (a) and auto-labelling pipeline (b).}
    \label{fig:drivepilot_autolabel}
\end{figure*}

Reliable perception is fundamental to AV deployment, supporting downstream prediction, planning, and control \cite{li2020comprehensive}. Modern systems typically employ multi-modal sensor suites combining LiDAR, RADAR, and cameras to leverage complementary sensing capabilities (Figure~\ref{fig:MB_DrivePilot}). While LiDAR provides accurate geometric information, RADAR offers robust velocity estimates under adverse conditions, and cameras supply rich semantic context, each modality exhibits inherent limitations when used in isolation \cite{yeong2021sensor}. Robust sensor fusion is therefore essential for comprehensive scene understanding \cite{arnold2020survey}.

A critical challenge in perception is accurate object geometry estimation. Many tracking methods approximate traffic participants as point targets, despite their spatial extent and multi-measurement nature. Extended Object Tracking (EOT) addresses this limitation by jointly estimating kinematics and shape \cite{koch2016tracking}, which is particularly important in dense urban environments with vulnerable road users. Classical EOT approaches, such as random matrix models, provide efficient elliptical representations but struggle with occlusions and complex object geometries \cite{feldmann2010tracking, MBTargetSig1}. Learning-based methods offer greater flexibility by inferring shape directly from sensor data \cite{meyer2020learning}, yet face challenges related to data annotation, generalization, and robustness under sparse or noisy measurements \cite{wang2021pointaugmenting}.

Graph Neural Networks (GNNs) have recently shown promise in modeling spatial and temporal dependencies in structured perception data \cite{wang2019dynamic}, as demonstrated by geometry-aware trackers such as 3DMOTFormer \cite{3dmotformer-iccv}. However, while large-scale datasets including KITTI \cite{geiger2013vision}, nuScenes \cite{caesar2020nuscenes}, and Waymo \cite{sun2020scalability} have enabled such advances, production systems often operate under strict computational and bandwidth constraints and expose only object-level tracks rather than raw sensor measurements \cite{duraisamy2013track}. These constraints motivate the development of data- and compute-efficient shape estimation methods suitable for real-world deployment.
To address these challenges, this work introduces the \textbf{L}earned \textbf{E}xtension of \textbf{O}bjects (LEO) framework for production-oriented extended object tracking. The key contributions are:

\begin{itemize}[topsep=0pt, partopsep=0pt, itemsep=1pt, parsep=0pt]
    \item A spatio-temporal architecture that leverages Graph Attention Network (GAT) blocks, originally proposed by \cite{veličković2018graph}, to enable adaptive shape estimation under production constraints.
    \item A parallelogram-based ground-truth formulation that generalizes bounding geometries to represent both rectangular and articulated objects such as trucks with trailers.
    \item A dual-attention mechanism that jointly captures intra-modal temporal dynamics and inter-modal spatial dependencies across multi-sensor tracks for robust fusion and sequential learning.
    \item Comprehensive evaluation on large-scale, real-world automotive datasets, demonstrating accurate, and computationally efficient performance across diverse driving scenarios.
\end{itemize}

\section{Related Works}
\label{sec:related_works}

\paragraph{Deep Learning for Object Detection}

The advent of deep learning has enabled models to learn complex geometric representations directly from multi-modal datasets with ground-truth 3D annotations. Early CNN-based approaches, such as PointPillars and SECOND \citep{lang2019pointpillars,yan2018second}, process voxelized inputs to produce oriented bounding boxes efficiently, while point-based methods like PointNet++ \citep{qi2017pointnet++} operate directly on raw point clouds. Transformer-based architectures, including DETR3D and BEVFormer \citep{wang2022detr3d,li2024bevformer}, exploit attention in Bird's-Eye View representations. Multi-modal fusion strategies, e.g., camera-LiDAR-RADAR integration \citep{yeong2021sensor,bai2022transfusion}, further enhance robustness under challenging conditions. Recent end-to-end EOT frameworks, such as CenterTrack \citep{zhou2020tracking}, TrackFormer \citep{meinhardt2022trackformer}, and TransTrack \citep{sun2020transtrack}, integrate detection, association, and shape estimation in a unified pipeline. By leveraging temporal embeddings and attention mechanisms, these models maintain object identities and consistent shape estimates across frames, even under occlusions or missed detections.

\paragraph{Extended Object Tracking}
Classical Extended Object Tracking (EOT) used Bayesian filters with simple parametric shapes like ellipses \citep{feldmann2010tracking,lan2019extended}, offering efficiency but limited expressiveness. Learning-based tracking integrates temporal consistency via transformers, exemplified by TrackFormer and TransTrack \citep{meinhardt2022trackformer,sun2020transtrack}. However, there is not much literature on deep-learned extension of objects in the context of EOT.

\section{Geometric Method and Auto-Labeling}
\label{sec:geometric_method}

\begin{figure*}[htbp]
\centering
\begin{subfigure}[t]{0.40\textwidth}
    \centering
    \includegraphics[width=\linewidth]{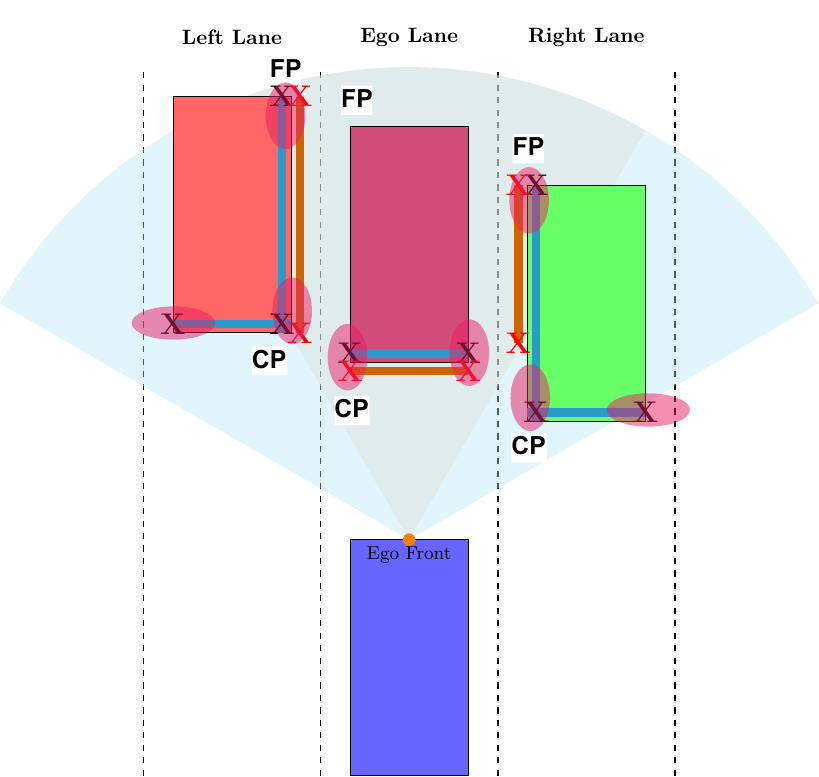}
    \caption{Sensor target shape types}
    \label{fig:object_types}
\end{subfigure}
\hspace{1.5cm}
\begin{subfigure}[t]{0.40\textwidth}
    \centering
    \includegraphics[width=\textwidth, trim=570 0 0 0, clip]{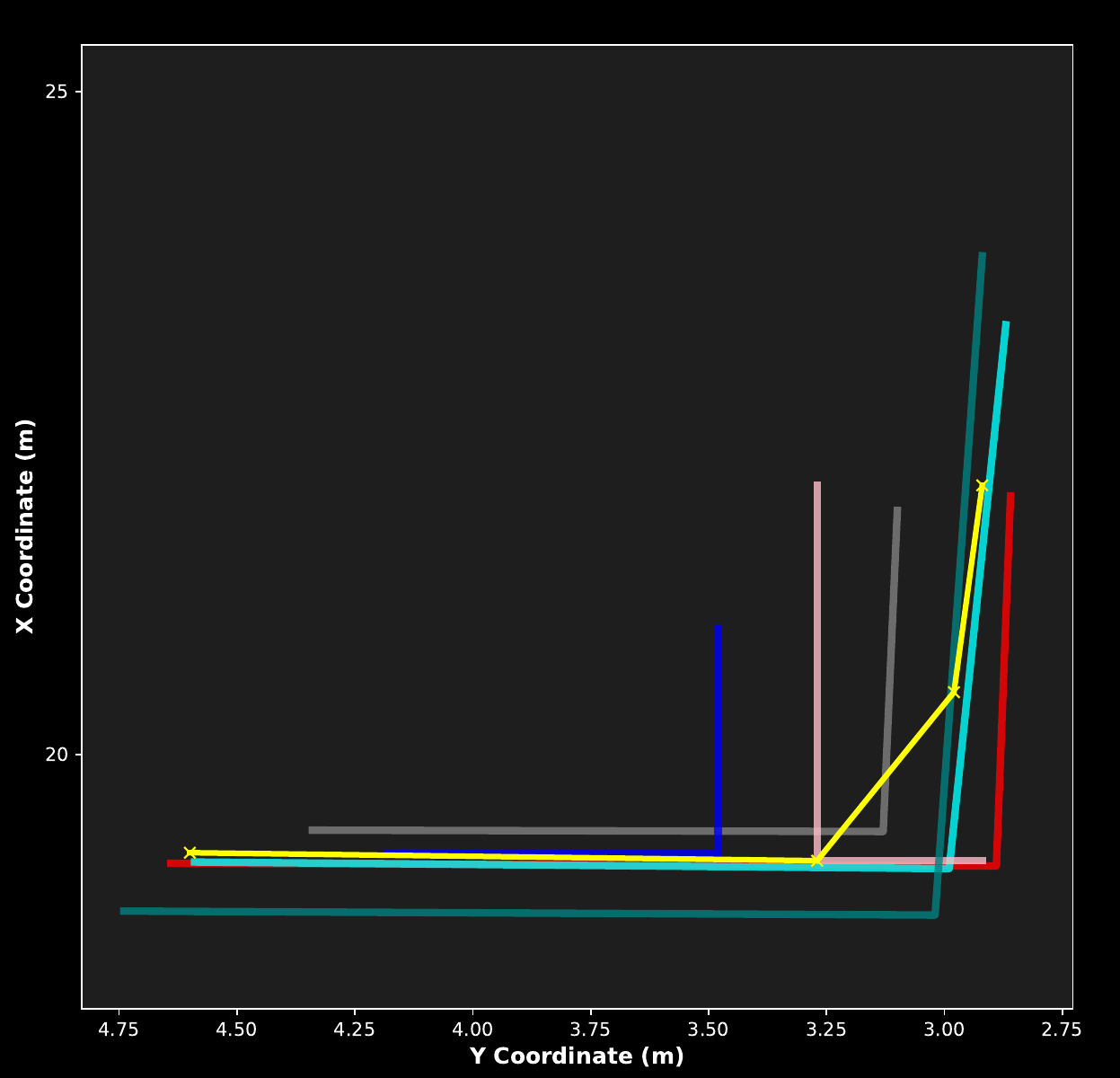}
    \caption{Target as detected by multiple sensors}
    \label{fig:single_target_inputs}
\end{subfigure}
\caption{Comparison of sensor field-of-view–based shape abstractions under occlusions, with overlays of the FOVs for \textcolor{orange}{RADAR} (60°) and \textcolor{cyan}{LiDAR} (120°), along with oval-shaped extension point covariances and lane-wise points for evaluation (a) and multi-sensor target shape segments (b).}
\label{fig:combined}
\end{figure*}

In series-production vehicles, raw sensor data is typically unavailable due to bandwidth, certification, and proprietary constraints from suppliers, resulting in perception modules that output high-level object tracks rather than low-level measurements \citep{duraisamy2013track}. These sensor tracks contain kinematic estimates, classification attributes, state covariances, and coarse object extents, abstracting away raw point clouds or image detections. \citep{CombiTor1} presents combination of this information granularity \citep{Fusion2023Patent} to achieve improved data association and fusion quality. Track-level fusion has emerged as a practical paradigm~\cite{bar2001estimation, tian2012track}, enabling modular integration of sensors and robustness across automotive platforms. Each sensor delivers object hypotheses in the form 
\begin{equation}
List_{\text{sens}} = \{\hat{\mathbf{x}}_i, \mathbf{P}_i, Ext_j\}
\label{eq:eot_hypo}
\end{equation}
where $\hat{\mathbf{x}}_i$ is the estimated kinematic state, $\mathbf{P}_i$ the covariance, and $Ext_j$ the $j$-th extension point with $m \leq 3$ depending on sensor resolution. The fusion task defines a function that generates a consistent fused representation of objects in Equation~\ref{eq:fusedshape}.
\begin{equation}
FusedShape = f(\hat{x}_{sens,i}, P_{sens,i})
\label{eq:fusedshape}
\end{equation}

Objects are abstracted as primitive geometric types depending on sensor modality and resolution depicted in Figure~\ref{fig:object_types}: \textbf{L-shapes} for high-resolution sensors like LiDAR capturing both edges and object in sensor's FOV, \textbf{I-shapes} when only one edge is visible, such as vehicle in front of ego vehicle or occluded, and \textbf{point-shapes} typical of RADAR with limited resolution at far ranges. This representation enables handling heterogeneity and partial observability across modalities. The hybrid fusion framework is modular, comprising kinematic state fusion with Kalman Filter (KF) or Covariance Intersection (CI), and shape extension fusion using computational geometry~\citep{duraisamy2016track, Fusion2023Patent}. Segment association relies on spatial and orientation criteria, using Hausdorff distance with threshold $d_{\text{Hausdorff}} < 2~\si{\meter}$ and angular constraint $\theta < 30^\circ$.
\begin{equation}
d_{Hausdorff} = \max\left( d(S_1, S_2), d(S_2, S_1) \right)
\label{eq:Hausdorff_dist}
\end{equation}

Once the association is established, segment endpoints are confidence-weighted inversely with their covariance determinant
\begin{equation}
Weight \propto \frac{1}{|\Sigma|}
\label{eq:invdet_weighting}
\end{equation}
prioritizing high-certainty observations. To conservatively combine correlated sensor data, Covariance Intersection (CI) is used, e.g.,
\begin{equation}
\sum_{FusionStart}^{-1} = \omega \sum_{S1start}^{-1} + (1-\omega) \sum_{S2start}^{-1}
\label{eq:fusion_start}
\end{equation}

with $\omega \in [0,1]$ balancing uncertainty contributions. Experimental validation on a Mercedes-Benz prototype with RADAR, LiDAR, and stereo cameras demonstrated sub-\SI{10}{\centi\metre} lateral accuracy, full modularity at the track level, and industrial readiness, highlighting the suitability of track-level fusion for safety-certified automotive perception stacks. In continuation of this model-based approach, the fused object shapes having three extension points serve as reliable auto-labels, fused tracks, that are subsequently utilized to supervise the training of LEO \citep{haag2020baas}. As illustrated in Figure~\ref{fig:autolabelling}, this establishes a closed-loop framework where geometric fusion not only enables modular perception in production systems but also provides consistent training targets for data-driven methods, thereby bridging model-based and learning-based paradigms within the automotive perception stack.

\begin{figure*}[htbp]
    \centering
    \includegraphics[trim=0 330 0 30,clip,width=1\textwidth]{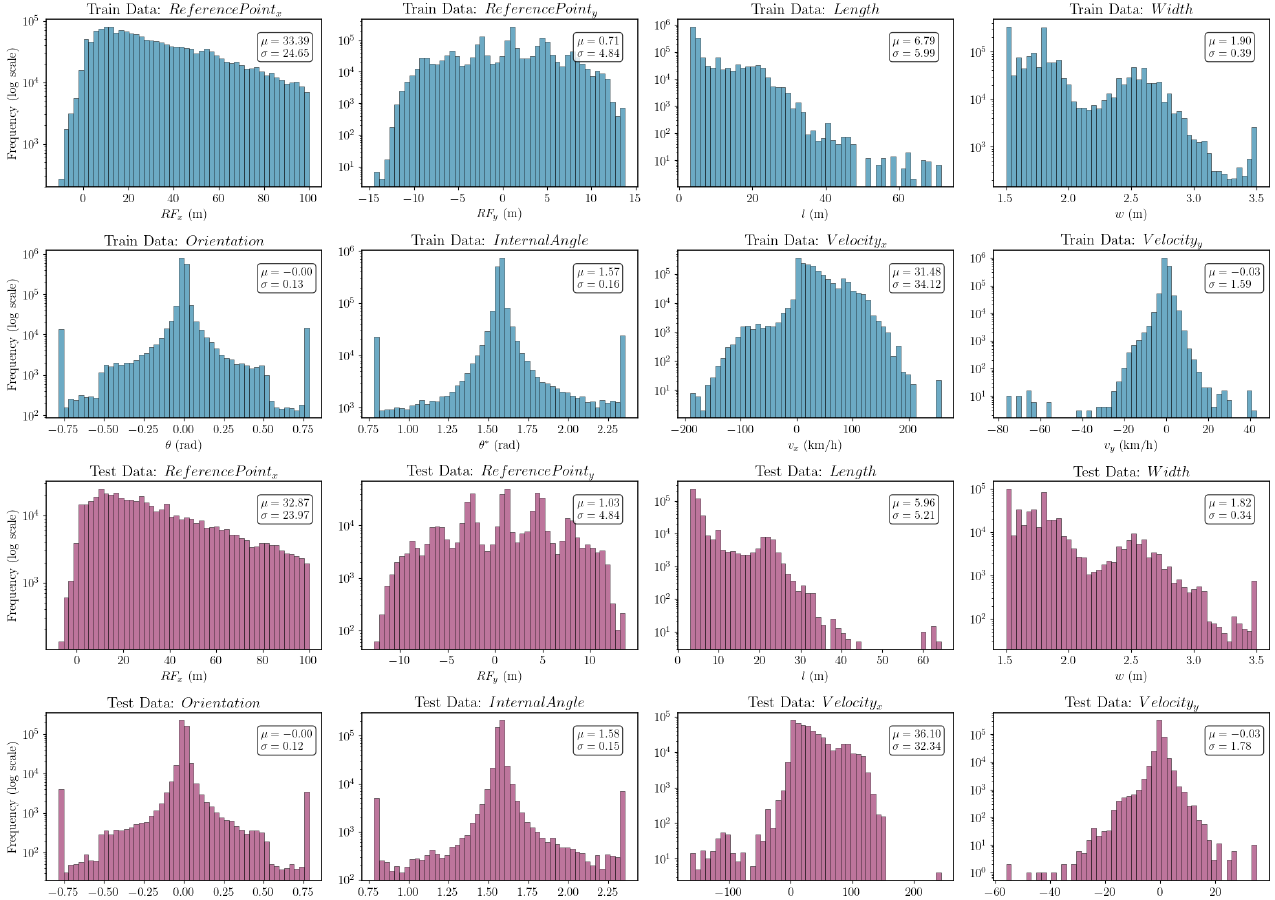}
    \caption{Statistics of target vehicle characteristics in train and test sequences}
    \label{fig:label_stats}
\end{figure*}

\section{LEO: Graph Attention Network Based Shape Estimation}
\label{sec:gat_model}

\begin{figure*}[htbp]
    \centering
    \includegraphics[trim=50 267 300 113,clip,width=1\textwidth]{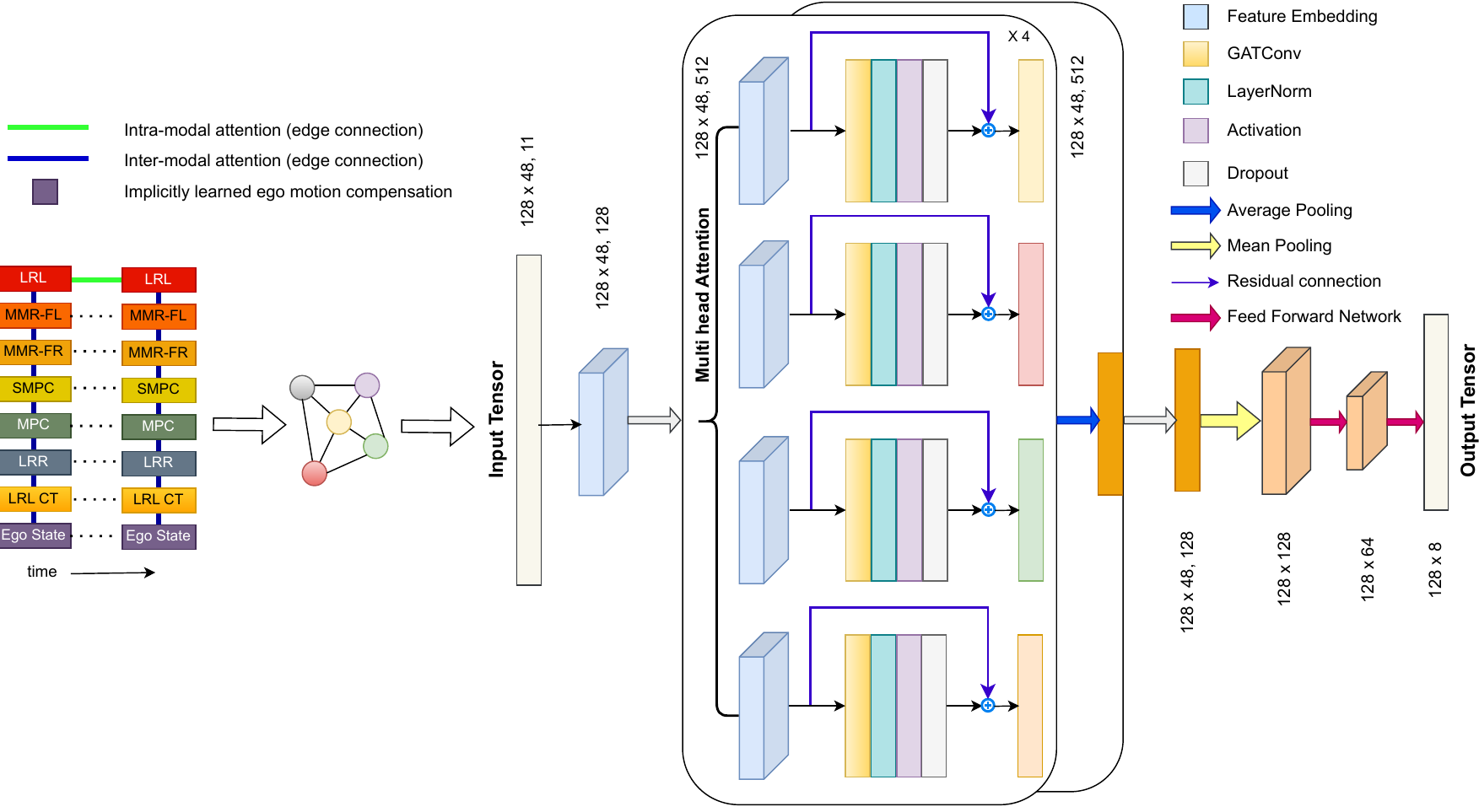}
    \caption{LEO architecture : Tracks from multi-modality sensors are first embedded with state vectors and timestamps, concatenated across six frames (120\,ms), and represented as a spatio-temporal graph with intra- and inter-modal edges for GAT-based attention. The LEO architecture then projects inputs ($128 \times 48 \times 11$) into a latent space ($128 \times 48 \times 128$), processes them through four stacked GATConv \cite{veličković2018graph} layers with dual attention, normalization, ELU activation, dropout, and residual connections, and aggregates multi-head outputs via pooling into $128 \times 128$ embeddings. A final feed-forward projection maps these to $128 \times 8$ parallelogram parameters $\hat{\mathbf{y}}$, enabling efficient joint spatio-temporal reasoning for shape fusion.}
    \label{fig:gat_architecture}
\end{figure*}

\paragraph{Parallelogram-Based Object Representation}
Traditional rectangular bounding boxes inadequately capture articulated or disjoint geometries, such as trucks with trailers. Since the sensor tracks in our dataset do not impose right-angle constraints, we represent objects as parallelograms, where the fourth vertex is obtained by completing the shape from three ordered extension points of the fused objects from geometric fusion. Each object is parameterized by its left rear vertex (reference point: $RF_x, RF_y$), dimensions ($l, w$), orientation and internal angle ($\theta, \theta^*$), and velocities ($v_x, v_y$), following the DIN~70000 standard~\citep{haken2015grundlagen}. This formulation generalizes rectangular cases ($\theta^* = 90^\circ$) while accommodating complex geometries through flexible angular constraints, as illustrated in Figure~\ref{fig:parallelogram_label}. The resulting state vector or label is:
\begin{equation}
\hat{\mathbf{y}} = \{RF_x, RF_y, l, w, \theta, \theta^*, v_x, v_y\} \in \mathbb{R}^8
\end{equation}

\subsection{Problem Formulation and Graph Construction :}
\label{subsec:problem_formulation}

\begin{figure*}[htbp]
    \centering
    \begin{subfigure}[t]{0.40\textwidth}
        \centering
        \includegraphics[trim=3cm 11.5cm 8cm 5cm, clip, width=\textwidth]{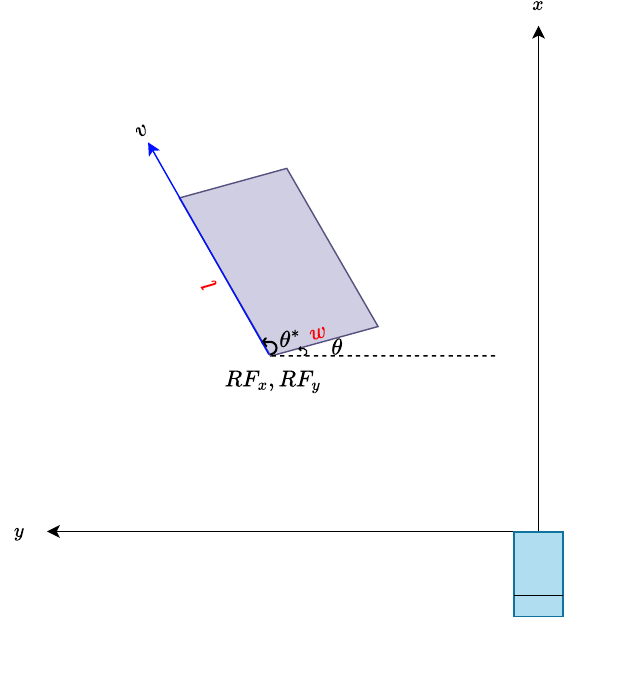}
        \caption{Parallelogram-shaped object representation in the ego coordinate frame.}
        \label{fig:parallelogram_label}
    \end{subfigure}%
    \hfill
    \begin{subfigure}[t]{0.60\textwidth}
        \centering
        \includegraphics[trim=2cm 10cm 1cm 3.5cm, clip, width=\textwidth]{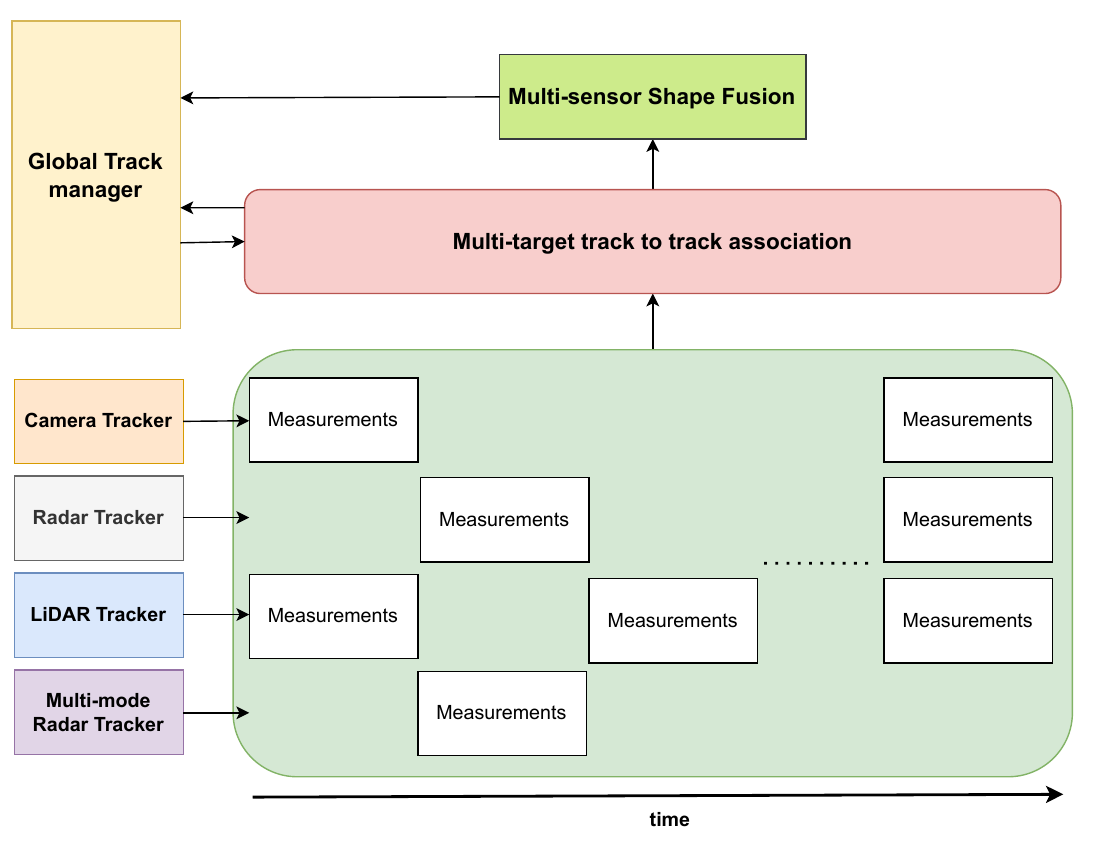}
        \caption{Shape Fusion Architecture}
        \label{fig:shapefusion_architecture}
    \end{subfigure}
    \caption{Parallelogram object representation with velocity vector represented as an \textcolor{blue}{arrow} from the reference point, which is at the left-rear vertex (a) and the proposed Shape Fusion architecture (b).}
    \label{fig:combined_figures}
\end{figure*}

We formulate multi-modal sensor fusion as a spatio-temporal graph learning problem \cite{fey2019fast} over heterogeneous sensor measurements with varying sampling rates as illustrated in Figure~\ref{fig:shapefusion_architecture}. The temporal alignment pipeline processes raw measurements from RADAR (\SI{60}{\hertz}), LiDAR (\SI{40}{\hertz}), and cameras (\SI{80}{\hertz}) through dedicated trackers, synchronizing outputs in \SI{20}{\milli\second} intervals within a \SI{120}{\milli\second} sliding window, producing target states and extension points (Figure~\ref{fig:single_target_inputs}). As sensors fire asynchronously at different frequencies, missing detections at a given timestamp are handled by propagating the most recent measurement in the data stream. Shape cues, primarily from LiDAR contours, are abstracted into L-shapes using geometric feature extraction and a dual-line RANSAC procedure~\citep{ling2024ransac} for robustness against outliers.

The spatio-temporal graph $\mathcal{G} = (\mathcal{V}, \mathcal{E})$ comprises $48$ nodes: $6$ ego-motion nodes encoding velocity, yaw rate, acceleration, and timestamp and $42$ sensor nodes from seven modalities (\textbf{L}ong-\textbf{R}ange \textbf{L}iDAR, \textbf{L}ong-\textbf{R}ange \textbf{R}ADAR, \textbf{M}ulti-\textbf{M}ode \textbf{R}ADAR \textbf{F}ront \textbf{R}ight, \textbf{M}ulti-\textbf{M}ode \textbf{R}ADAR \textbf{F}ront \textbf{L}eft, \textbf{M}ulti-\textbf{P}urpose \textbf{C}amera, \textbf{L}iDAR contour, and \textbf{S}tereo \textbf{M}ulti-\textbf{P}urpose \textbf{C}amera) across six timestamps. Each sensor node $\mathbf{n}_{s,t-k}$ encodes an 11-dimensional feature vector:
\begin{equation}
\mathbf{f}^{(t-k)} = [x_1, x_2, x_3, y_1, y_2, y_3, \sigma_x^2, \sigma_y^2, v_x, v_y, \Delta t]^T
\label{eq:node_features}
\end{equation}
representing extension points $x_i, y_i$, uncertainties $\sigma_x, \sigma_y$, velocities $v_x, v_y$, and temporal offset $\Delta t$ in seconds to the fusion timestamp. Ego-motion nodes are similarly encoded as
\begin{equation}
\mathbf{n}_{\text{ego},t-k} = [v_{t-k}, \dot{\psi}_{t-k}, a_{t-k}, \ldots, \Delta t_k]^T \in \mathbb{R}^{11},
\end{equation}
allowing implicit learning of ego-motion compensation. The edge set $\mathcal{E}$ captures temporal evolution and cross-modal dependencies through three edge types:
\begin{align}
\mathcal{E}_{\mathrm{temporal}} &= \{(\mathbf{n}_{s,t-k}, \mathbf{n}_{s,t-(k-1)}) \mid s \in [1,8],\, k \in [1,5]\} \\
\mathcal{E}_{\mathrm{spatial}} &= \{(\mathbf{n}_{s_i,t-k}, \mathbf{n}_{s_j,t-k}) \mid s_i \neq s_j,\, k \in [0,5]\} \\
\mathcal{E}_{\mathrm{self}} &= \{(\mathbf{n}_{s,t-k}, \mathbf{n}_{s,t-k}) \mid s \in [1,8],\, k \in [0,5]\}
\end{align}

\subsection{Dual Attention Mechanism, Training and Network Architecture}
\label{subsec:dual_attention}

LEO employs a dual-attention mechanism (Figure~\ref{fig:gat_architecture}) that independently models temporal consistency, i.e., shape evolution and motion dynamics, within individual sensor modalities (intra-modal), while simultaneously integrating complementary spatial information across modalities (inter-modal) \citep{veličković2018graph}. The resulting unified attention formulation is given by:

\begin{equation}
\alpha_{ij}^{(m)} = 
\frac{\exp\!\Big(\mathrm{LeakyReLU}\!\big(\mathbf{a}_{m}^{\top} 
[\mathbf{W}_{m}\mathbf{h}_i \,\|\, \mathbf{W}_{m}\mathbf{h}_j]\big)\Big)}
{\sum\limits_{k \in \mathcal{N}_i^{(m)}} \exp\!\Big(\mathrm{LeakyReLU}\!\big(\mathbf{a}_{m}^{\top} 
[\mathbf{W}_{m}\mathbf{h}_i \,\|\, \mathbf{W}_{m}\mathbf{h}_k]\big)\Big)}
\label{eq:modular_attention}
\end{equation}
where $m$ denotes the attention type: $\mathrm{intra}$ corresponds to temporal neighbors $\mathcal{N}_i^{\mathrm{temporal}}$, capturing motion-consistent patterns, while $\mathrm{inter}$ corresponds to spatial neighbors $\mathcal{N}_i^{\mathrm{spatial}}$, aggregating complementary information across modalities. $\mathbf{W}_m$ and $\mathbf{a}_m$ are the learnable weight and attention query vectors for the respective modality.

The final attention coefficients balance temporal and spatial contributions:
\begin{equation}
\alpha_{ij}^{\text{st}} = \lambda \cdot \alpha_{ij}^{\text{intra}} + (1-\lambda) \cdot \alpha_{ij}^{\text{inter}}
\end{equation}
enabling adaptive weighting based on data availability and quality. Message passing follows:
\begin{equation}
\mathbf{h}_i^{(l+1)} = \sigma\!\left(\sum_{j \in \mathcal{N}_i} \alpha_{ij}^{\text{st}} \mathbf{W}^{(l)} \mathbf{h}_j^{(l)}\right)
\end{equation}

\paragraph{Training Objective and Optimization}
The training objective combines parameter-level regression with geometry-aware supervision through a composite loss function:
\begin{equation}
\mathcal{L}_{\text{total}} = \mathcal{L}_{\text{param}} + \lambda_{\text{IoU}} \mathcal{L}_{\text{IoU}}
\end{equation}

The parameter loss applies SmoothL1 regression to individual components:
\begin{equation}
\mathcal{L}_{\text{param}} = \sum_{i \in \{RF_x, RF_y, l, w, \theta, \theta^*, v_x, v_y\}} \beta_i \cdot \text{SmoothL1}(\hat{\mathbf{y}}_i, \mathbf{y}_i)
\end{equation}
where $\beta_i$ weights balance parameter importance based on estimation difficulty and downstream impact. The geometry loss combines Generalized IoU \cite{rezatofighi2019generalized} and Distance IoU \cite{zheng2020distance} to enforce spatial consistency:
\begin{equation}
\mathcal{L}_{\text{IoU'}} = \alpha \cdot \mathcal{L}_{\text{GIoU}} + (1 - \alpha) \cdot \mathcal{L}_{\text{DIoU}}
\end{equation}

where GIoU ensures enclosure constraints while DIoU enforces centroid alignment. Training is conducted using the Adam optimizer \citep{kingma2015adam} with an initial learning rate of $1 \times 10^{-4}$ and plateau-based decay (factor $0.75$). The loss function uses $\beta = 1$ and $\alpha = 0.5$. The model is trained for up to $50$ epochs with a batch size of $128$ and gradient clipping at a norm of $3.0$. Early stopping with a patience of $5$ epochs is applied to prevent overfitting, with convergence typically achieved around $40$ epochs, beyond which validation performance stagnates.

\section{Evaluation}
\label{sec:evaluation}

\paragraph{Dataset Description} The proposed model is evaluated on proprietary data from the Mercedes-Benz SAE Level-3 DRIVE PILOT system, comprising multi-sensor fusion outputs of static and dynamic objects with ego vehicle states across diverse US and European driving environments. The dataset includes \SI{12.3}{\hour} of training and \SI{2.31}{\hour} of testing data, covering highway driving and controlled \emph{cut-in} sequences to enrich safety-critical scenarios (Table~\ref{tab:dataset}). Traffic participants range from passenger cars and commercial vehicles to articulated trucks and vulnerable road users, spanning $RF_x \in [-10,100]\,\si{m}$, $RF_y \in [-12,12]\,\si{m}$, and dimensions from compact cars ($\approx 3\,\si{m}$) to articulated trucks ($>70\,\si{m}$). Ego states include velocities up to $140\,\si{km/h}$, yaw rates $\pm0.6\,\si{rad/s}$, and accelerations $-10$ to $+5\,\si{m/s^2}$, capturing both longitudinal motion and lateral maneuvers, providing a production-relevant benchmark for common and safety-critical events as in Figure~\ref{fig:label_stats}.

\begin{table}[h]
    \centering
    \small
    \setlength{\tabcolsep}{4pt}
    \caption{Dataset composition for training and testing sequences.}
    \label{tab:dataset}
    \begin{tabular}{lcccc}
        \toprule
        & Driving & Cut-Ins & Hours & Fusion Objects \\
        \midrule
        Train Sequence & 326 & 410 & 12.3 hrs & 1.46 mil. \\
        Test Sequence  & 79  & 60  & 2.31 hrs & 0.44 mil. \\
        \bottomrule
    \end{tabular}
\end{table}

\subsection{Evaluation Strategy}
\label{subsec:eval_strategy}

\begin{table*}[htbp]
\centering
\caption{Global KPIs for Shape Estimation of Fused Objects for LEO}
\label{tab:combined_kpi_gat}
\small
\begin{tabular}{l|cc|cc|cc|cc}
\toprule
\multirow{3}{*}{\textbf{Parameter}} 
& \multicolumn{4}{c|}{\textbf{$l_1$}} 
& \multicolumn{4}{c}{\textbf{$l_2$}} \\
\cmidrule(lr){2-5} \cmidrule(lr){6-9}
& \multicolumn{2}{c|}{\textbf{with $\alpha_{ij}^{\text{inter}}$}}
& \multicolumn{2}{c|}{\textbf{without $\alpha_{ij}^{\text{inter}}$}} 
& \multicolumn{2}{c|}{\textbf{with $\alpha_{ij}^{\text{inter}}$}} 
& \multicolumn{2}{c}{\textbf{without $\alpha_{ij}^{\text{inter}}$}} \\
& \text{MAE} & \text{Error (\%)} 
& \text{MAE} & \text{Error (\%)} 
& \text{MAE} & \text{Error (\%)} 
& \text{MAE} & \text{Error (\%)} \\
\midrule
GIoU (-)          & 0.78 & --    & 0.27 & --    & 0.76 & --    & 0.31 & -- \\
DIoU (-)          & 0.82 & --    & 0.23 & --    & 0.76 & --    & 0.34 & -- \\
$RF_x$ (m)        & 0.21 & 0.60  & 3.98 & 11.45 & 0.40 & 1.35  & 3.18 & 10.49 \\
$RF_y$ (m)        & 0.11 & 2.94  & 0.64 & 16.42 & 0.14 & 4.95  & 0.67 & 23.67 \\
$l$ (m)           & 0.43 & 10.16 & 3.55 & 83.29 & 2.22 & 11.62 & 8.60 & 44.88 \\
$w$ (m)           & 0.08 & 4.88  & 0.37 & 21.03 & 0.12 & 5.22  & 0.27 & 11.30 \\
$\theta$ (rad)    & 0.04 & --    & 0.13 & -- & 0.05 & --    & 0.11 & -- \\
$\theta^*$ (rad)  & 0.05 & 3.09  & 0.13 & 8.13   & 0.05 & 3.24  & 0.11 & 6.98 \\
$v_x$ (m/s)       & 0.24 & 2.01  & 3.37 & 28.15  & 0.30 & 3.27  & 2.73 & 29.07 \\
$v_y$ (m/s)       & 0.10 & --    & 0.23 & --  & 0.12 & --    & 0.21 & -- \\
\bottomrule
\end{tabular}
\end{table*}

Evaluation is conducted on the complete test dataset, using region-based overlaps of oriented parallelograms (GIoU and DIoU) and the Mean Absolute Error (MAE) of output parameters. Objects are stratified by length, with $l_1 \in [3,10]\,\si{m}$ representing cars and light commercial vans, and $l_2 > 10\,\si{m}$ representing buses, trucks and trailers. The evaluation is reported along three complementary axes: first, \textit{global performance} across all objects in the ROI, providing an overall benchmark of model robustness; second, evaluation on \textit{open source dataset} and third, a \textit{lane-wise analysis}, where results are partitioned by object centroid position into ego lane (EL: $[-1.5,\,1.5]\,\si{m}$), left lane (LL: $(1.5,\,4.5]\,\si{m}$), and right lane (RL: $[-4.5,\,-1.5)\,\si{m}$). This structure ensures that both aggregate accuracy and spatially resolved safety-critical contexts for motion planning are systematically assessed.

\paragraph{Global Performance}
LEO achieves high spatial accuracy with GIoU/DIoU scores of $0.76$--$0.82$ across both object categories. Reference point estimation remains below $0.4\,\si{m}$ (MAE) with relative errors under $5\%$, while dimensional accuracy is consistent: car-sized objects ($l_1$) attain MAE of $0.43\,\si{m}$ in length and $0.08\,\si{m}$ in width, and articulated objects ($l_2$) reach $2.22\,\si{m}$ and $0.12\,\si{m}$, corresponding to $10$--$12\%$ relative errors. Orientation errors remain below $3^\circ$ and velocity estimates are precise within $0.3\,\si{m/s}$ ($<1.3\,\si{km/h}$). Implemented in PyTorch and benchmarked on an RTX $2080$~Ti GPU with an $18$-core CPU, LEO processes samples at avg. inference time $\sim 13.5\,\si{ms}$ (runtime $\sim 30\,\text{FPS}$) with minimal memory usage (\SI{0.02}{\gibi\byte}), demonstrating robust, and computationally efficient performance suitable for real-time deployment after appropriate optimization.

\paragraph{Evaluation on Public Dataset}
\begin{table}[t]
\centering
\small
\setlength{\tabcolsep}{4pt}
\renewcommand{\arraystretch}{0.95}
\caption{Class-wise performance comparison for LiDAR-only (CenterPoint \citep{yin2021center}) and LiDAR+RADAR (LEO) for test split on VoD dataset.}
\label{tab:classwise_cp_leo}
\begin{tabular}{llccccc}
\toprule
\textbf{Class} & \textbf{Cfg} &
\textbf{\shortstack{Recall\\(IoU=0.25)}} & \textbf{mASE} & \textbf{mAOE} & \textbf{mAVE} \\
\midrule
\multirow{2}{*}{Car}
& L   & 0.75 & 0.40 & 0.60 & 1.27 \\
&L+R & \textbf{0.81} & \textbf{0.16} & \textbf{0.58} & \textbf{0.45} \\
\midrule
\multirow{2}{*}{Pedestrian}
&L   & 0.73 & \textbf{0.20} & 0.85 & 1.01 \\
&L+R & \textbf{0.75} & 0.33 & \textbf{0.58} & \textbf{0.51} \\
\midrule
\multirow{2}{*}{Cyclist}
 & L   & 0.32 & 0.66 & 0.93 & 4.17 \\
 & L+R & \textbf{0.64} & \textbf{0.26} & \textbf{0.39} & \textbf{1.14} \\
\bottomrule
\end{tabular}
\end{table}

\begin{table}[htbp]
\centering
\small
\setlength{\tabcolsep}{3.5pt}
\caption{\textbf{Ablation Study}: Lane-wise performance analysis for LEO without LRR, LRL, and SMPC. For each setting, results are reported across three consecutive rows corresponding to the ego lane (EL), left lane (LL), and right lane (RL).}
\label{tab:mae_lane_results_gat_extended_lrr_ablated}
\begin{tabular}{lccccc}
\toprule
\textbf{GIoU} & \textbf{CP\textsubscript{x}} & \textbf{CP\textsubscript{y}} & \textbf{FP\textsubscript{x}} & \textbf{FP\textsubscript{y}} \\
\midrule
\textbf{0.55} / \textbf{0.22} & 1.39 / 0.36 & 0.28 / 0.16 & 2.37 / 13.39 & 0.35 / 0.49 \\
0.76 / \textbf{0.48} & 0.48 / 0.43 & 0.22 / 1.30 & 1.02 / \textbf{7.54} & 0.28 / 0.58 \\
0.75 / \textbf{0.47} & 0.52 / 0.64 & 0.11 / 0.12 & 1.11 / \textbf{7.21} & 0.16 / 0.34 \\
0.89 / 0.83 & 0.33 / 0.35 & 0.10 / 0.18 & 0.45 / 1.12 & 0.13 / 0.43 \\
0.76 / 0.73 & \textbf{0.69} / 0.39 & \textbf{0.30} / 0.31 & \textbf{1.27} / 2.64 & 0.37 / 0.46 \\
0.74 / 0.64 & \textbf{0.93} / 0.82 & \textbf{0.17} / 0.17 & \textbf{1.60} / 3.53 & 0.23 / 0.35 \\
0.89 / 0.83 & 0.25 / 0.33 & 0.09 / 0.21 & 0.36 / 1.04 & 0.12 / 0.51 \\
0.75 / 0.71 & 0.43 / \textbf{0.45} & 0.26 / 0.28 & 0.95 / 2.91 & 0.31 / 0.49 \\
0.75 / 0.66 & 0.64 / 0.70 & 0.15 / 0.18 & 1.24 / 3.46 & 0.22 / \textbf{0.39} \\
\bottomrule
\end{tabular}
\end{table}

\begin{table}[htbp]
\centering
\small
\setlength{\tabcolsep}{3pt}
\caption{Lane-wise analysis for LEO. Each row corresponds to a lane (EL, LL, RL), and values report mean IoU (--) and MAE, grouped by length classification ($l_1/l_2$).}
\label{tab:mae_lane_results_gat_extended}
\begin{tabular}{lccccc}
\toprule
\textbf{GIoU} & \textbf{CP\textsubscript{x}} & \textbf{CP\textsubscript{y}} & \textbf{FP\textsubscript{x}} & \textbf{FP\textsubscript{y}} \\
\midrule
0.91 / 0.84 & 0.10 / 0.27 & 0.07 / 0.16 & 0.21 / 0.87 & 0.10 / 0.37 \\
0.79 / 0.77 & 0.19 / 0.34 & 0.20 / 0.25 & 0.64 / 2.30 & 0.23 / 0.40 \\
0.77 / 0.71 & 0.23 / 0.55 & 0.10 / 0.13 & 0.82 / 3.17 & 0.17 / 0.31 \\
\bottomrule
\end{tabular}
\end{table}

The View-of-Delft (VoD) dataset~\citep{palffy2022multi} provides calibrated and time-synchronized data from LiDAR, RGB cameras, and 4D RADAR, captured in urban traffic scenarios in Delft, the Netherlands. It comprises 8,693 annotated frames and includes vulnerable road users such as pedestrians and cyclists, making it suitable for evaluating 3D object detection and shape estimation methods \citep{deng2024robust}. Public datasets such as nuScenes, Waymo, or KITTI primarily provide raw sensor measurements (e.g., point clouds or images) and do not expose the heterogeneous, asynchronous \emph{track-level} representations (e.g., RADAR covariance ellipses, geometric LiDAR contours, or stereo tracks) required by the proposed architecture. To enable evaluation on a public benchmark while remaining consistent with the track-level formulation used in the proprietary dataset, the VoD dataset is employed with a controlled track-generation pipeline.

Specifically, LiDAR detections are generated using \textbf{CenterPoint}~\citep{yin2021center}, while RADAR returns are associated to fused tracks via \textit{adaptive clustering}~\citep{haag2019extended}, followed by L-shape fitting to extract corner features~\citep{ling2024ransac}. These track-level representations are then used as input to the graph-based network, retraining the model for sensor-level fusion. VoD provides 100\,ms synchronized frames, and a sliding window of three frames is used to aggregate detections before fusion. Table~\ref{tab:classwise_cp_leo} reports class-wise evaluation results using metrics: \textit{Recall (IoU=0.25)}, \textit{mASE}, \textit{mAOE}, and \textit{mAVE}. Compared to the LiDAR-only baseline, LEO with L+R consistently improves recall and reduces scale, orientation, and velocity errors, particularly for cars, pedestrians and cyclists, demonstrating robust performance even without vision input.

\begin{figure*}[htbp]
    \centering
    \begin{subfigure}[b]{0.49\textwidth}
        \centering
        \includegraphics[trim=40 0 20 0, clip, width=\textwidth]{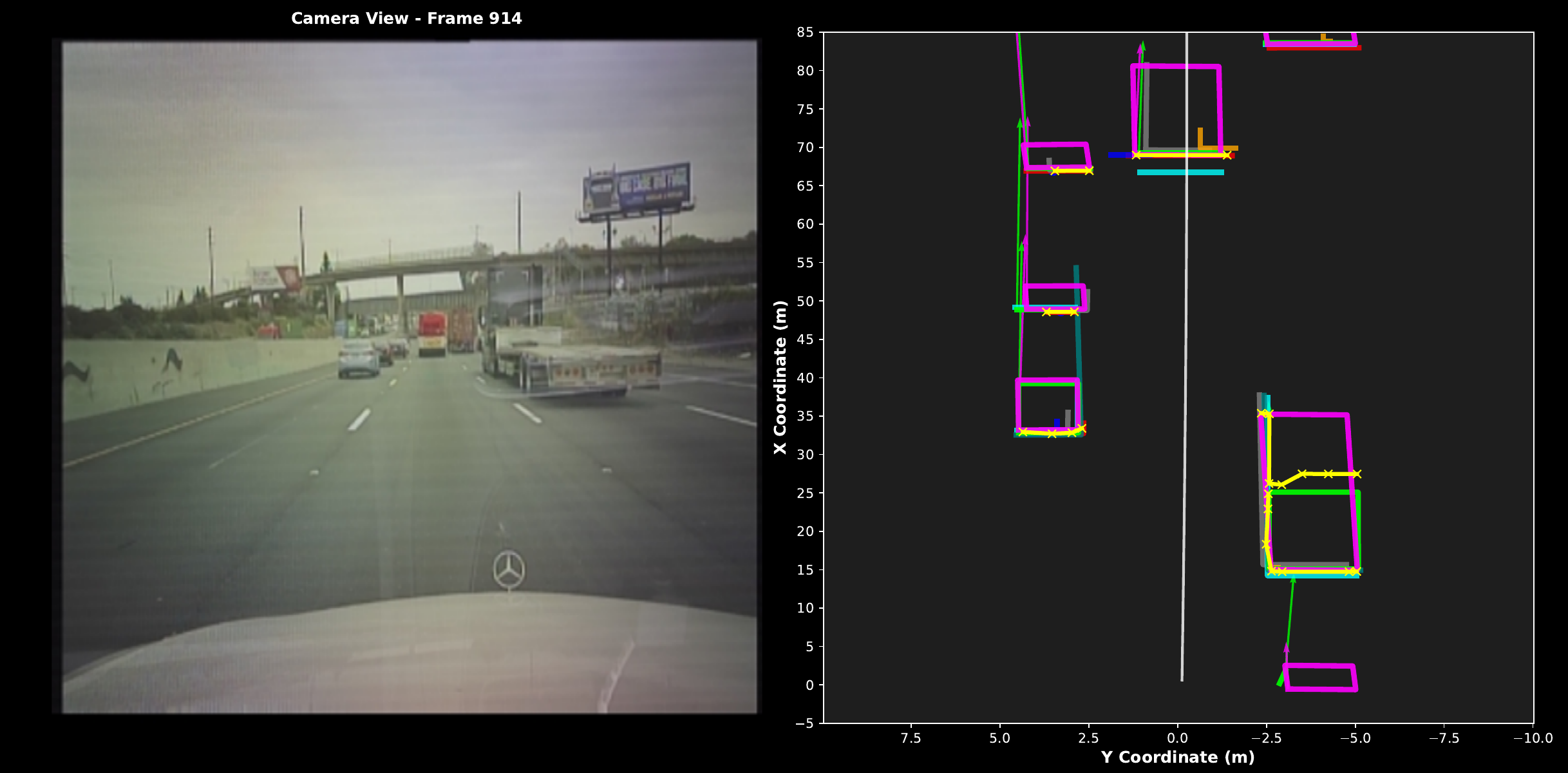}
        \caption{Highway Driving with occluded targets}
        \label{fig:sub_eval_results}
    \end{subfigure}
    \hfill
    \begin{subfigure}[b]{0.49\textwidth}
        \centering
        \includegraphics[trim=40 0 20 0, clip, width=\textwidth]{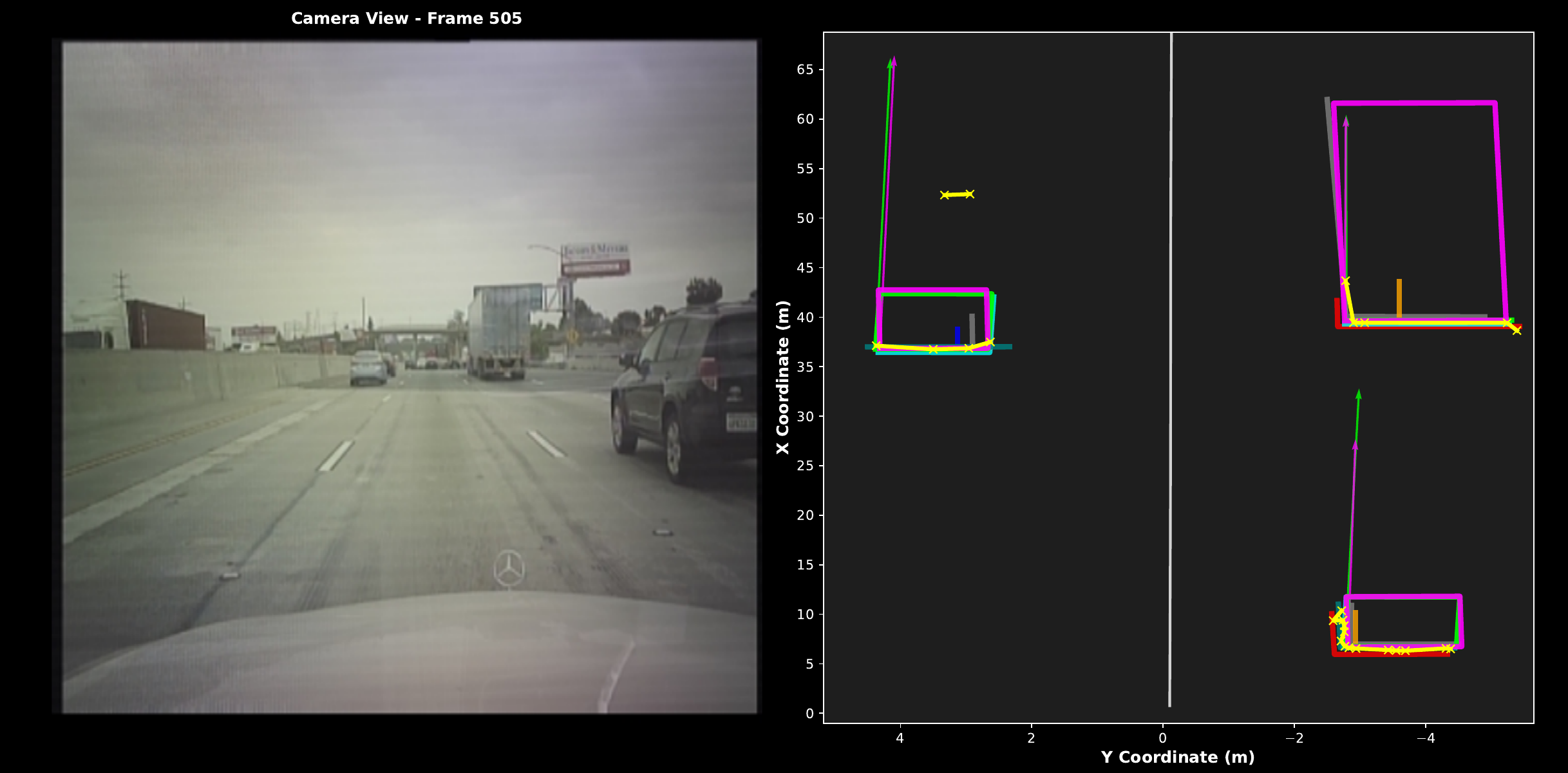}
        \caption{Unoccluded objects}
        \label{fig:sub_good_results}
    \end{subfigure}

    \vspace{0.3cm}

    \begin{subfigure}[b]{0.49\textwidth}
        \centering
        \includegraphics[trim=40 0 20 0, clip, width=\textwidth]{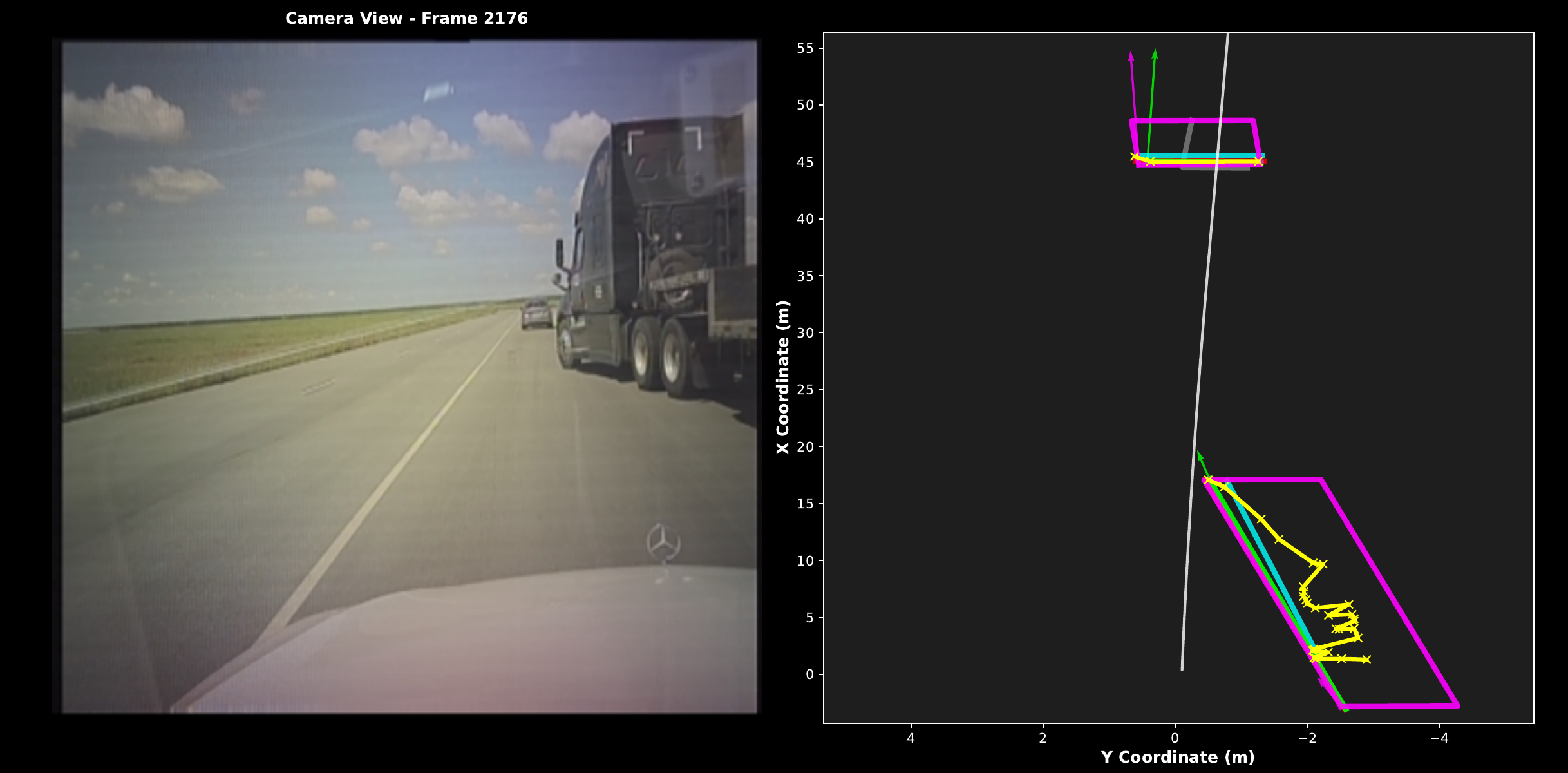}
        \caption{Articulated Vehicle entering into ego lane}
        \label{fig:sub_cutin}
    \end{subfigure}
    \hfill
    \begin{subfigure}[b]{0.49\textwidth}
        \centering
        \includegraphics[trim=40 0 20 0, clip, width=\textwidth]{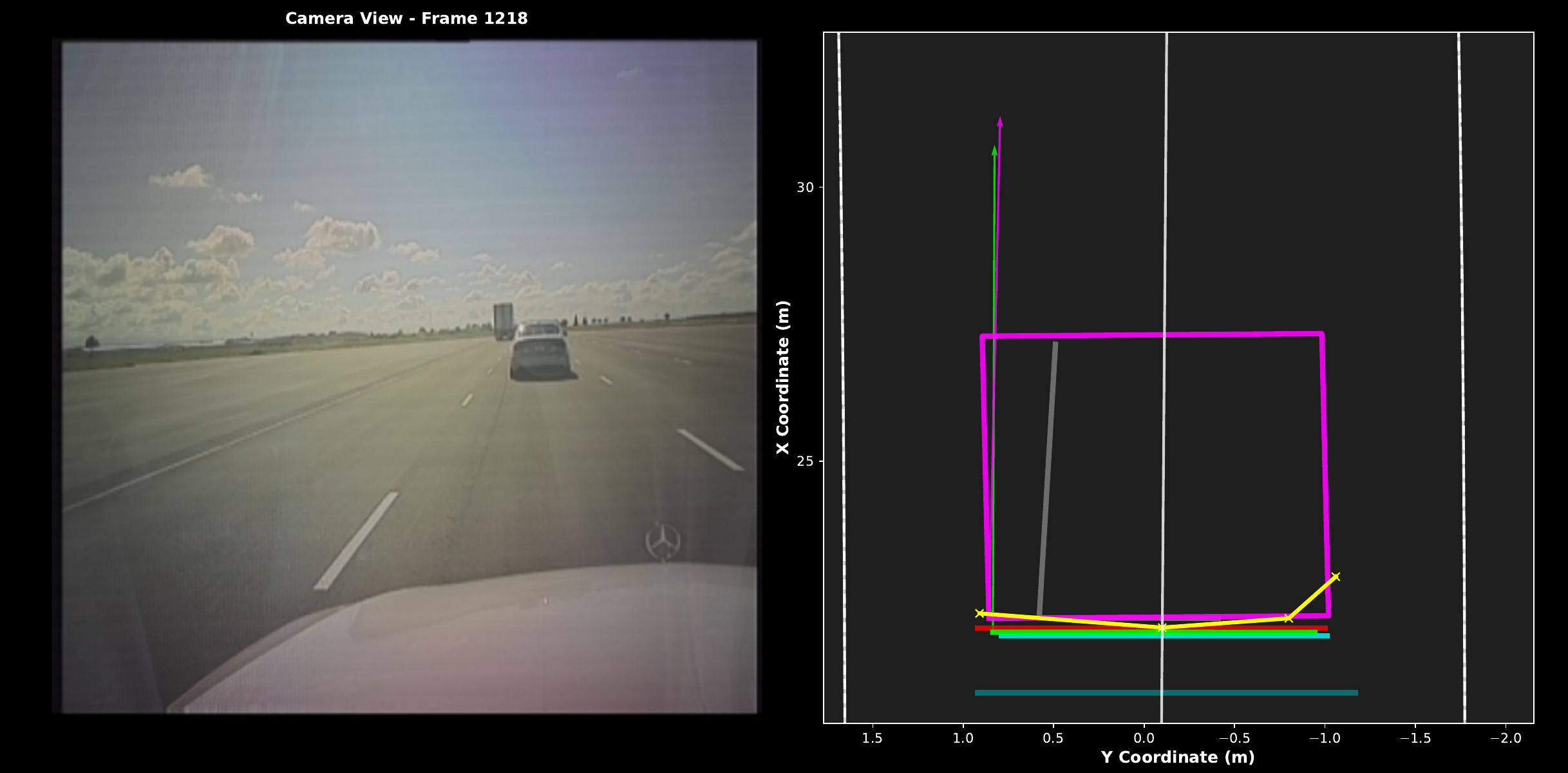}
        \caption{Lead vehicle in ego lane}
        \label{fig:sub_ego}
    \end{subfigure}
    \caption{Evaluation results across diverse real world scenarios.}
    \label{fig:evaluation_combined}
\end{figure*}

\paragraph{Lane-wise Performance}
Table~\ref{tab:mae_lane_results_gat_extended} presents lane-wise performance of LEO. In the ego lane, the model achieves the highest accuracy, with GIoU above $0.9$ for $l_1$ and $0.84$ for $l_2$, and ($10$--$27\,\si{cm}$) CP errors, corresponding to the lead vehicle directly ahead of the ego car. This is attributed to favorable sensor coverage and consistent rear-edge visibility of lead vehicles, enabling precise learning of dimensions and orientation. In adjacent lanes, performance degrades moderately (GIoU $0.77$–$0.79$), as sensor placement, FOV, and resolution cause different object edges to be visible for different sensors with varying covariances of extension points for each track. The adaptive fusion mechanism compensates for these differences by weighting inputs through graph attention, yielding robust estimates. Notably, $l_2$ show larger farthest-point errors ($2$--$3\,\si{m}$), yet the overall high GIoU across lanes substantiates the effectiveness of the proposed approach in handling heterogeneous observability while prioritizing safety-critical objects in the ego lane.

\subsection{Ablation Study}
\label{subsec:ablation_lanewise}

LEO's dual-attention mechanism, intra-modal attention for temporal consistency and inter-modal attention $\alpha_{ij}^{\mathrm{inter}}$ for cross-sensor spatial fusion provides a structured interpretation of the degradation patterns. Removing LRR (Table~\ref{tab:mae_lane_results_gat_extended_lrr_ablated}) yields the most severe collapse, particularly in the ego lane where only the rear edge of the lead vehicle is typically visible. RADAR’s longitudinal penetrability and returns from within the vehicle body supply depth cues that LiDAR and SMPC cannot recover under occlusion; without these signals, intra-modal attention loses its primary constraint on object extent, causing GIoU to drop to $0.55/0.22$ and $\text{FP}_x$ to explode to $13.39\,\si{m}$. LRL ablation produces a different failure mode: its long-range precision and visibility into adjacent lanes are critical for inter-modal spatial aggregation, and removing it markedly increases $\text{CP}_x$ and $\text{FP}_x$ (up to $1.60\,\si{m}$), especially under cross-lane occlusions. Thus, removing the LRL destabilizes the stability of the reference point crucial for object tracking and geometric extent.

SMPC ablation produces the mildest degradation: the stereo module primarily contributes near-range depth cues and contour precision, so its removal leads to slight increases in reference-point and lateral errors while largely preserving global object geometry. In contrast, disabling inter-modal attention (Table~\ref{tab:combined_kpi_gat}) highlights its essential role in maintaining global multi-sensor consistency. Without cross-sensor relational weighting, reference-point drift increases sharply ($RF_x$: $0.21 \rightarrow 3.98\,\si{m}$) and dimensional accuracy deteriorates substantially ($l$: $0.43 \rightarrow 3.55\,\si{m}$, $+83\%$), even though temporal attention remains active. These results underscore the benefit of jointly modelling spatial, temporal, and uncertainty-aware cues within the attention mechanism. Overall, LRR is indispensable for stable longitudinal extent inference under occlusion especially for articulated vehicles. LRL supports cross-lane robustness, reference-point stability, and spatial completeness, whereas SMPC serves as a geometric refinement layer. Finally, inter-modal attention remains fundamental for ensuring globally coherent and uncertainty-consistent multi-sensor shape estimation.

\subsection{Qualitative Analysis}
\label{subsec:gat_results_lanes}

Figure~\ref{fig:evaluation_combined} illustrates a qualitative evaluation of LEO across highway and proving ground scenarios. Learned shapes (\textcolor{magenta}{magenta}) are compared with model-based fusion outputs (\textcolor{green}{green}), while sensor tracks from individual modalities are shown in additional colors with velocity vectors as arrows. In highway driving (Figure~\ref{fig:sub_eval_results}), input tracks often exhibit shortened bounding box lengths under sparse observations, particularly for distant vehicles. LEO adapts to these degraded inputs while maintaining consistent geometry, and suppresses spurious SMPC detections that erroneously merge multiple objects into one through attention weighting. For articulated objects such as a truck–trailer in the right lane, the model accurately reconstructs the full extent by combining LiDAR contours with near-range SMPC depth cues, outperforming rule-based fusion which systematically underestimates length. In unoccluded cases (Figure~\ref{fig:sub_good_results}), orientation and dimensions align closely with sensor inputs. During dynamic maneuvers such as cut-ins (articulated vehicle merging into ego lane) and emergency braking (Figure~\ref{fig:sub_cutin}), the model produces temporally stable predictions by integrating long-range RADAR length cues with multi-modal velocity estimates, which is critical for safe planning. For near-field targets (Figure~\ref{fig:sub_ego}), depth inconsistencies across modalities are resolved by prioritizing high-confidence LiDAR contours, yielding corrected and reliable shape estimates.

\section{Conclusion And Future Work}
\label{sec:conclusion}
This work presented \textbf{LEO}, a spatio-temporal GAT-based framework for adaptive shape estimation in extended object tracking on track-level sensor inputs. Leveraging a parallelogram-based ground truth, LEO models both rectangular and articulated target geometries, while its dual-attention mechanism enables joint reasoning over temporal dynamics and spatial dependencies for robust multi-sensor fusion. Extensive evaluation on large-scale real-world datasets confirms that LEO delivers accurate, stable, and computationally efficient shape representations across diverse scenarios, validating its suitability for deployment. Future work will explore uncertainty-aware estimation, domain adaptation, continual learning, and lightweight variants for embedded platforms, as well as integration with planning modules to quantify the impact of shape-aware perception on automated driving safety and efficiency.

\bibliography{example_paper}{}

@inproceedings{haag2020baas,
  title={Baas: Bayesian tracking and fusion assisted object annotation of radar sensor data for artificial intelligence application},
  author={Haag, Stefan and Duraisamy, Bharanidhar and Govaers, Felix and Koch, Wolfgang and Fritzsche, Martin and Dickmann, J{\"u}rgen},
  booktitle={2020 IEEE Radar Conference (RadarConf20)},
  pages={1--6},
  year={2020},
  organization={IEEE}
}

@inproceedings{haag2019extended,
  title={Extended object tracking assisted adaptive clustering for radar in autonomous driving applications},
  author={Haag, Stefan and Duraisamy, Bharanidhar and Govaers, Felix and Koch, Wolfgang and Fritzsche, Martin and Dickmann, J{\"u}rgen},
  booktitle={2019 Sensor Data Fusion: Trends, Solutions, Applications (SDF)},
  pages={1--7},
  year={2019},
  organization={IEEE}
}

@inproceedings{deng2024robust,
  title={Robust 3d object detection from lidar-radar point clouds via cross-modal feature augmentation},
  author={Deng, Jianning and Chan, Gabriel and Zhong, Hantao and Lu, Chris Xiaoxuan},
  booktitle={2024 IEEE International Conference on Robotics and Automation (ICRA)},
  pages={6585--6591},
  year={2024},
  organization={IEEE}
}

@inproceedings{
veličković2018graph,
title={Graph Attention Networks},
author={Petar Veličković and Guillem Cucurull and Arantxa Casanova and Adriana Romero and Pietro Liò and Yoshua Bengio},
booktitle={International Conference on Learning Representations},
year={2018},
url={https://openreview.net/forum?id=rJXMpikCZ},
}

@inproceedings{
kingma2015adam,
title={Adam: A Method for Stochastic Optimization},
author={Diederik P. Kingma, Jimmy Ba},
booktitle={3rd International Conference for Learning Representations, San Diego, 2015},
year={2015},
url={https://arxiv.org/pdf/1412.6980},
}

@book{haken2015grundlagen,
  title={Grundlagen der Kraftfahrzeugtechnik},
  author={Haken, Karl-Ludwig},
  year={2015},
  publisher={Carl Hanser Verlag GmbH Co KG}
}

@article{feldmann2010tracking,
  title={Tracking of extended objects and group targets using random matrices},
  author={Feldmann, Michael and Fr{\"a}nken, Dietrich and Koch, Wolfgang},
  journal={IEEE Transactions on Signal Processing},
  volume={59},
  number={4},
  pages={1409--1420},
  year={2010},
  publisher={IEEE}
}

@article{geiger2013vision,
  title={Vision meets robotics: The kitti dataset},
  author={Geiger, Andreas and Lenz, Philip and Stiller, Christoph and Urtasun, Raquel},
  journal={The international journal of robotics research},
  volume={32},
  number={11},
  pages={1231--1237},
  year={2013},
  publisher={Sage Publications Sage UK: London, England}
}

@inproceedings{caesar2020nuscenes,
  title={nuscenes: A multimodal dataset for autonomous driving},
  author={Caesar, Holger and Bankiti, Varun and Lang, Alex H and Vora, Sourabh and Liong, Venice Erin and Xu, Qiang and Krishnan, Anush and Pan, Yu and Baldan, Giancarlo and Beijbom, Oscar},
  booktitle={Proceedings of the IEEE/CVF conference on computer vision and pattern recognition},
  pages={11621--11631},
  year={2020}
}

@inproceedings{sun2020scalability,
  title={Scalability in perception for autonomous driving: Waymo open dataset},
  author={Sun, Pei and Kretzschmar, Henrik and Dotiwalla, Xerxes and Chouard, Aurelien and Patnaik, Vijaysai and Tsui, Paul and Guo, James and Zhou, Yin and Chai, Yuning and Caine, Benjamin and others},
  booktitle={Proceedings of the IEEE/CVF conference on computer vision and pattern recognition},
  pages={2446--2454},
  year={2020}
}

@article{wang2019dynamic,
  title={Dynamic graph cnn for learning on point clouds},
  author={Wang, Yue and Sun, Yongbin and Liu, Ziwei and Sarma, Sanjay E and Bronstein, Michael M and Solomon, Justin M},
  journal={ACM Transactions on Graphics (tog)},
  volume={38},
  number={5},
  pages={1--12},
  year={2019},
  publisher={Acm New York, NY, USA}
}

@book{koch2016tracking,
  title={Tracking and sensor data fusion},
  author={Koch, Wolfgang},
  year={2016},
  publisher={Springer}
}

@inproceedings{yin2021center,
  title={Center-based 3d object detection and tracking},
  author={Yin, Tianwei and Zhou, Xingyi and Krahenbuhl, Philipp},
  booktitle={Proceedings of the IEEE/CVF conference on computer vision and pattern recognition},
  pages={11784--11793},
  year={2021}
}

@INPROCEEDINGS{MBTargetSig1,
  author={Haag, Stefan and Duraisamy, Bharanidhar and Koch, Wolfgang and Dickmann, Jürgen},
  booktitle={2018 21st International Conference on Information Fusion (FUSION)}, 
  title={Radar and Lidar Target Signatures of Various Object Types and Evaluation of Extended Object Tracking Methods for Autonomous Driving Applications}, 
  year={2018},
  volume={},
  number={},
  pages={1746-1755},
  doi={10.23919/ICIF.2018.8455395}}

@misc{MB-DrivePilot2023-1,
	author = {Mercedes-Benz},
	title = {{M}ercedes-{B}enz DRIVE PILOT The front runner in automated driving and safety technologies.},
	howpublished = {\url{https://group.mercedes-benz.com/innovation/case/autonomous/drive-pilot-2.html}},
	year = {2023},
	note = {[Accessed 24-09-2025]},
}

@INPROCEEDINGS{CombiTor1,
  author={Duraisamy, Bharanidhar and Schwarz, Tilo and Wöhler, Christian},
  booktitle={2015 18th International Conference on Information Fusion (Fusion)}, 
  title={On track-to-track data association for automotive sensor fusion}, 
  year={2015},
  volume={},
  number={},
  pages={1213-1222},
  doi={}}

@article{qi2017pointnet++,
  title={Pointnet++: Deep hierarchical feature learning on point sets in a metric space},
  author={Qi, Charles Ruizhongtai and Yi, Li and Su, Hao and Guibas, Leonidas J},
  journal={Advances in neural information processing systems},
  volume={30},
  year={2017}
}

@inproceedings{lang2019pointpillars,
  title={Pointpillars: Fast encoders for object detection from point clouds},
  author={Lang, Alex H and Vora, Sourabh and Caesar, Holger and Zhou, Lubing and Yang, Jiong and Beijbom, Oscar},
  booktitle={Proceedings of the IEEE/CVF conference on computer vision and pattern recognition},
  pages={12697--12705},
  year={2019}
}

@inproceedings{meinhardt2022trackformer,
  title={Trackformer: Multi-object tracking with transformers},
  author={Meinhardt, Tim and Kirillov, Alexander and Leal-Taixe, Laura and Feichtenhofer, Christoph},
  booktitle={Proceedings of the IEEE/CVF conference on computer vision and pattern recognition},
  pages={8844--8854},
  year={2022}
}

@INPROCEEDINGS{3dmotformer-iccv,
  author={Ding, Shuxiao and Rehder, Eike and Schneider, Lukas and Cordts, Marius and Gall, Juergen},
  booktitle={2023 IEEE/CVF International Conference on Computer Vision (ICCV)}, 
  title={3DMOTFormer: Graph Transformer for Online 3D Multi-Object Tracking}, 
  year={2023},
  volume={},
  number={},
  pages={9750-9760},
  doi={10.1109/ICCV51070.2023.00897}}

@article{ling2024ransac,
  title={RANSAC-Based Planar Point Cloud Segmentation Enhanced by Normal Vector and Maximum Principal Curvature Clustering},
  author={Ling, Yibo and Wang, Yuli and Chan, Ting On},
  journal={ISPRS Annals of the Photogrammetry, Remote Sensing and Spatial Information Sciences},
  volume={10},
  pages={145--151},
  year={2024},
  publisher={Copernicus Publications G{\"o}ttingen, Germany}
}

@article{palffy2022multi,
  title={Multi-class road user detection with 3+ 1d radar in the view-of-delft dataset},
  author={Palffy, Andras and Pool, Ewoud and Baratam, Srimannarayana and Kooij, Julian FP and Gavrila, Dariu M},
  journal={IEEE Robotics and Automation Letters},
  volume={7},
  number={2},
  pages={4961--4968},
  year={2022},
  publisher={IEEE}
}

@article{yeong2021sensor,
  title={Sensor and sensor fusion technology in autonomous vehicles: A review},
  author={Yeong, De Jong and Velasco-Hernandez, Gustavo and Barry, John and Walsh, Joseph},
  journal={Sensors},
  volume={21},
  number={6},
  pages={2140},
  year={2021},
  publisher={MDPI}
}

@inproceedings{duraisamy2016track,
  title={Track level fusion of extended objects from heterogeneous sensors},
  author={Duraisamy, Bharanidhar and Gabb, Michael and Nair, Aswin Vijayamohnan and Schwarz, Tilo and Yuan, Ting},
  booktitle={2016 19th International Conference on Information Fusion (FUSION)},
  pages={876--885},
  year={2016},
  organization={IEEE}
}

@inproceedings{duraisamy2013track,
  title={Track level fusion algorithms for automotive safety applications},
  author={Duraisamy, Bharanidhar and Schwarz, Tilo and W{\"o}hler, Christian},
  booktitle={2013 International Conference on Signal Processing, Image Processing \& Pattern Recognition},
  pages={179--184},
  year={2013},
  organization={IEEE}
}

@inproceedings{tian2012track,
  title={Track-to-track fusion in linear and nonlinear systems},
  author={Tian, Xin and Yuan, Ting and Bar-Shalom, Yaakov},
  booktitle={Itzhack Y. Bar-Itzhack Memorial Symposium on Estimation, Navigation, and Spacecraft Control},
  pages={21--41},
  year={2012},
  organization={Springer}
}

@book{bar2001estimation,
  title={Estimation with applications to tracking and navigation: theory algorithms and software},
  author={Bar-Shalom, Yaakov and Li, X Rong and Kirubarajan, Thiagalingam},
  year={2001},
  publisher={John Wiley \& Sons}
}

@article{singh2015critical,
  title={Critical reasons for crashes investigated in the national motor vehicle crash causation survey},
  author={Singh, Santokh},
  journal={Traffic safety facts crash stats. Report No. DOT HS 812 115},
  year={2015},
  publisher={Washington, DC: National Highway Traffic Safety Administration}
}

@article{fagnant2015preparing,
  title={Preparing a nation for autonomous vehicles: opportunities, barriers and policy recommendations},
  author={Fagnant, Daniel J and Kockelman, Kara},
  journal={Transportation Research Part A: Policy and Practice},
  volume={77},
  pages={167--181},
  year={2015},
  publisher={Elsevier}
}

@article{yurtsever2020survey,
  title={A survey of autonomous driving: Common practices and emerging technologies},
  author={Yurtsever, Ekim and Lambert, Jacob and Carballo, Alexander and Takeda, Kazuya},
  journal={IEEE access},
  volume={8},
  pages={58443--58469},
  year={2020},
  publisher={IEEE}
}

@article{badue2021self,
  title={Self-driving cars: A survey},
  author={Badue, Claudine and Guidolini, R{\^a}nik and Carneiro, Raphael Vivacqua and Azevedo, Pedro and Cardoso, Vinicius B and Forechi, Avelino and Jesus, Luan and Berriel, Rodrigo and Paixao, Thiago M and Mutz, Filipe and others},
  journal={Expert Systems with Applications},
  volume={165},
  pages={113816},
  year={2021},
  publisher={Elsevier}
}

@article{li2020comprehensive,
  title={A comprehensive survey on the application of deep and reinforcement learning approaches in autonomous driving},
  author={Li, Yiming and Ibanez-Guzman, Javier},
  journal={Journal of Field Robotics},
  volume={37},
  number={5},
  pages={789--821},
  year={2020},
  publisher={Wiley Online Library}
}

@article{arnold2020survey,
  title={A survey on 3D object detection methods for autonomous driving applications},
  author={Arnold, Eduardo and Al-Jarrah, Omar Y and Dianati, Mehrdad and Fallah, Saber and Oxtoby, David and Mouzakitis, Alex},
  journal={IEEE Transactions on Intelligent Transportation Systems},
  volume={21},
  number={4},
  pages={1708--1733},
  year={2019},
  publisher={IEEE}
}

@inproceedings{meyer2020learning,
  title={Learning an uncertainty-aware object detector for autonomous driving},
  author={Meyer, Gregory P and Thakurdesai, Niranjan},
  booktitle={2020 IEEE/RSJ International Conference on Intelligent Robots and Systems (IROS)},
  pages={10521--10527},
  year={2020},
  organization={IEEE}
}

@inproceedings{wang2021pointaugmenting,
  title={Pointaugmenting: Cross-modal augmentation for 3d object detection},
  author={Wang, Chunwei and Ma, Chao and Zhu, Ming and Yang, Xiaokang},
  booktitle={Proceedings of the IEEE/CVF conference on computer vision and pattern recognition},
  pages={11794--11803},
  year={2021}
}

@inproceedings{rezatofighi2019generalized,
  title={Generalized intersection over union: A metric and a loss for bounding box regression},
  author={Rezatofighi, Hamid and Tsoi, Nathan and Gwak, JunYoung and Sadeghian, Amir and Reid, Ian and Savarese, Silvio},
  booktitle={Proceedings of the IEEE/CVF conference on computer vision and pattern recognition},
  pages={658--666},
  year={2019}
}

@inproceedings{zheng2020distance,
  title={Distance-IoU loss: Faster and better learning for bounding box regression},
  author={Zheng, Zhaohui and Wang, Ping and Liu, Wei and Li, Jinze and Ye, Rongguang and Ren, Dongwei},
  booktitle={Proceedings of the AAAI conference on artificial intelligence},
  pages={12993--13000},
  year={2020}
}

@inproceedings{bai2022transfusion,
  title={Transfusion: Robust lidar-camera fusion for 3d object detection with transformers},
  author={Bai, Xuyang and Hu, Zeyu and Zhu, Xinge and Huang, Qingqiu and Chen, Yilun and Fu, Hongbo and Tai, Chiew-Lan},
  booktitle={Proceedings of the IEEE/CVF conference on computer vision and pattern recognition},
  pages={1090--1099},
  year={2022}
}

@article{li2024bevformer,
  title={Bevformer: learning bird's-eye-view representation from lidar-camera via spatiotemporal transformers},
  author={Li, Zhiqi and Wang, Wenhai and Li, Hongyang and Xie, Enze and Sima, Chonghao and Lu, Tong and Yu, Qiao and Dai, Jifeng},
  journal={IEEE Transactions on Pattern Analysis and Machine Intelligence},
  year={2024},
  publisher={IEEE}
}

@inproceedings{wang2022detr3d,
  title={Detr3d: 3d object detection from multi-view images via 3d-to-2d queries},
  author={Wang, Yue and Guizilini, Vitor Campagnolo and Zhang, Tianyuan and Wang, Yilun and Zhao, Hang and Solomon, Justin},
  booktitle={Conference on Robot Learning},
  pages={180--191},
  year={2022},
  organization={PMLR}
}

@inproceedings{zhou2020tracking,
  title={Tracking objects as points},
  author={Zhou, Xingyi and Koltun, Vladlen and Kr{\"a}henb{\"u}hl, Philipp},
  booktitle={European conference on computer vision},
  pages={474--490},
  year={2020},
  organization={Springer}
}

@article{sun2020transtrack,
  title={Transtrack: Multiple object tracking with transformer},
  author={Sun, Peize and Cao, Jinkun and Jiang, Yi and Zhang, Rufeng and Xie, Enze and Yuan, Zehuan and Wang, Changhu and Luo, Ping},
  journal={arXiv preprint arXiv:2012.15460},
  year={2020}
}

@article{yan2018second,
  title={Second: Sparsely embedded convolutional detection},
  author={Yan, Yan and Mao, Yuxing and Li, Bo},
  journal={Sensors},
  volume={18},
  number={10},
  pages={3337},
  year={2018},
  publisher={Multidisciplinary Digital Publishing Institute}
}

@article{lan2019extended,
  title={Extended-object or group-target tracking using random matrix with nonlinear measurements},
  author={Lan, Jian and Li, X Rong},
  journal={IEEE Transactions on Signal Processing},
  volume={67},
  number={19},
  pages={5130--5142},
  year={2019},
  publisher={IEEE}
}

@article{fey2019fast,
  title={Fast graph representation learning with PyTorch Geometric},
  author={Fey, Matthias and Lenssen, Jan Eric},
  journal={arXiv preprint arXiv:1903.02428},
  year={2019}
}

@misc{Fusion2023Patent,
  author = {Duraisamy, Bharanidhar and Haag, Stefan and Fritzsche, Martin.},
  title = {Verfahren zu einer hybriden Multisensor-Fusion für automatisiert betriebene Fahrzeuge, DE102023001184A1},
  number = {DE102023001184A1},
  year = {2023},
  month = {March},
  day = {27}
}
\bibliographystyle{plainnat}

\end{document}